\documentclass[a4paper,12pt,onecolumn]{article}

\usepackage{FAS}
\usepackage[utf8]{inputenc}
\usepackage[english]{babel}
\usepackage{csquotes}
\usepackage{graphicx}
\usepackage{xcolor}
\usepackage{array}
\usepackage{caption}
\usepackage[inkscapearea=page]{svg}
\usepackage{float}
\usepackage{tcolorbox}
\usepackage{booktabs}
\usepackage{tabularx} 
\usepackage{adjustbox}
\usepackage[
  backend=biber,
  style=ieee,
  citestyle=ieee
]{biblatex}
\begin{document}
%
\title{Towards Safer Cities: A Design Process for Perception Datasets in Urban Driving Scenarios}
%
%
\author{Alexander Kroth
\thanks{All authors are with the Institute of Automotive Engineering Darmstadt (FZD), TU Darmstadt, Otto-Berndt-Str. 2, 64287 Darmstadt, Germany (e-mail: \{alexander.kroth; cristian.martinez; steven.peters\}@tu-darmstadt.de).}
, Cristian Martinez%
\footnotemark[1]
\ and Steven Peters%
\footnotemark[1]
%
}
%
%
\date{}

\maketitle \thispagestyle{empty}

\begin{abstract}
Urban environments present significant challenges for the perception systems of automated driving due to their high complexity, characterized by dense traffic, diverse obstacles, or a wide range of possible scenarios. While numerous automotive datasets exist, few focus on urban settings, and the majority  are acquired exclusively on passenger cars and overlook other transport modalities such as heavy-duty trucks, busses and trams. In contrast to passenger vehicles, these platforms operate under different geographical and operational constraints; including traversing pedestrian zones or platform-specific tasks such as movement of trailer or winter service. In this work a method is proposed for a guided dataset creation based on an analysis of existing datasets using user-generated map data. Furthermore, representative truck-specific use cases are identified and exemplified to demonstrate the applicability of the proposed methodology and to support targeted acquisition planning.
\end{abstract}


\section{Introduction} \label{introduction}
Heavy-duty vehicles, buses and trams form the backbone of urban transportation for both goods and passengers. However, their operation is threatened by an already existing and growing shortage of qualified drivers~\cite{ji-hyland_what_2022, hr_wie_nodate}. Long-haul applications such as main-line rail or trucking are facing similar problems, but are already in the focus of both research and industry~\cite{schuster_will_2023}. To tackle this issue automation is proposed as a possible solution. In main-line rail, this is reflected by the introduction of the European Train Control System (ETCS), which is already being tested in selected projects~\cite{european_commission_etcs_nodate}. In the trucking sector, companies such as Torc  and Aurora are pursuing to solve the hub-to-hub use-case, focused on highway operation~\cite{torc_robotics_solutions_2023, aurora_aurora_nodate}. 

In contrast, great progress has been made in the research and adoption of automated driving in passenger cars. This is illustrated by both the large amount of available datasets and research based on them in this sector~\cite{liu_survey_2024}. Furthermore, commercial services such as Waymo currently operate autonomous taxi fleets in multiple cities in the United States and are expected to expand to London in early 2026~\cite{waymo_hello_nodate}. Datasets form the basis for the development, training, and evaluation of both modular and end-to-end trained automated driving stacks. The scope of existing datasets closely aligns with automated driving solutions we see today.

This work demonstrates that current datasets only partially cover the requirements of truck and tram automation in urban environments. Based on an analysis of existing datasets and user-generated map data, relevant truck- and tram-specific scenarios are identified and used to derive Operational Design Domains (ODDs) for both platforms. Building on these findings, a methodology for targeted data acquisition is proposed to systematically address the identified gaps and support the creation of future urban truck and tram datasets.

\let\clearpage\relax
\section{Methodology} 
\label{chapter:methodology}
To identify missing pieces in existing datasets for urban scenarios it is first necessary to find a fitting taxonomy to describe those datasets. For this purpose, the concept of an Operational Design Domain (ODD) as defined by Scholtes et al. \cite{scholtes_6-layer_2021} in the six layer model is adopted. The six layer model mostly focuses on the external world, omitting the ego-vehicle. While this is valid within the stated purpose of ODD description for a vehicle, when describing a dataset, especially one that also should be usable for closed-loop training of End-To-End Networks, the vehicle capabilities and scenarios present in the dataset also come into play. To address this limitation, the ODD definition proposed by Mendiboure et al.~\cite{mendiboure_operational_2023} is additionally considered, as it incorporates vehicle capabilities and scenario descriptions. Consequently, the six-layer model extended by the concepts introduced by Mendiboure et al. is used as the basis for describing, analyzing, and comparing datasets throughout this work.

When a dataset is recorded, all elements captured by the sensor setup become part of the recorded ODD and can be categorized according to the six-layer model. Since the sensor configuration determines which parts of the surrounding environment can be perceived, it is also considered part of the dataset description. Compared to passenger cars, both trucks and trams allow sensors to be mounted at higher positions, which can improve the overview of the surrounding scene. At the same time, these different viewpoints may introduce a domain shift and reduce the transferability of perception algorithms trained on passenger-car datasets. Consequently, vehicle-specific characteristics and sensor configurations are relevant components of the ODD. Finally, the scenarios present in the recordings must also be considered, as identical entities can have different relevance depending on their interactions and spatial relationships within a scene.

With the recording of a dataset, all elements captured by the sensor setup become part of the recorded ODD and can be categorized according to the six layer model. Since the sensor configuration determines which parts of the surrounding environment can be perceived, it is also considered part of the dataset description. ICompared to passenger cars, both trucks and trams allow sensors to be mounted at higher positions, which can improve the overview of the surrounding scene. At the same time, these different viewpoints may introduce a domain shift and reduce the transferability of perception algorithms trained on passenger-car datasets. Therefore, vehicle capabilities are also a relevant component of the ODD. Finally, the scenarios present in the recordings must also be considered, as identical entities can have different relevance depending on their interactions and spatial relationships within a scene.

Not all ODD elements are equally accessible within existing datasets. Following the six layer model, the accessibility of information contained in datasets is evaluated, and where information is not directly available, methods for its retrospective acquisition are investigated. Two considered external data sources are Open-Meteo \cite{Zippenfenig_Open-Meteo} for weather data and OpenStreetMap (OSM) \cite{OpenStreetMap} for map information. Both services can provide valuable complementary information and enable a more comprehensive characterization of a dataset's ODD. However, the resulting information may be less accurate than data obtained directly from the dataset itself. OSM relies on user-generated content and may therefore contain inconsistencies or outdated annotations. Similarly, Open-Meteo derives historical weather information from meteorological models, which may not fully reflect the exact conditions experienced by the vehicle at the time and location of data acquisition.

\begin{figure}[hptb]
    \centering{\includegraphics[width=1\textwidth]{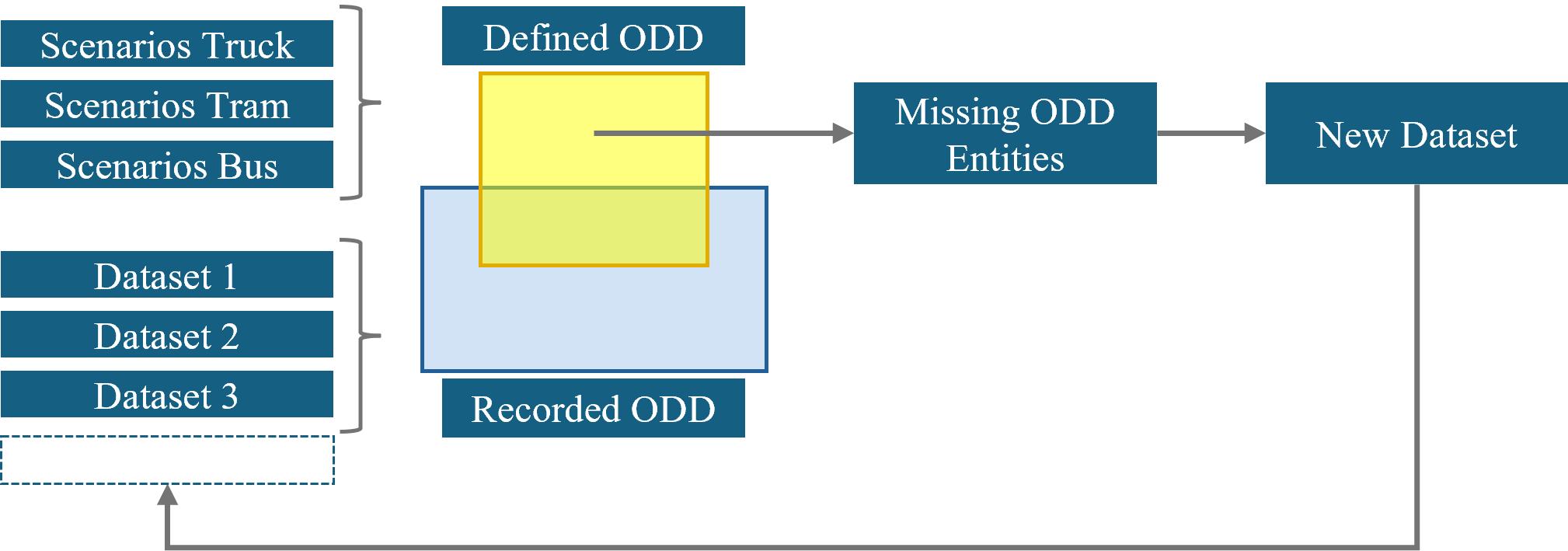}}
    \caption{Acquisition strategy is informed by missing ODD entities acquired by comparing defined ODD and recorded ODD}
    \label{fig:methodology}
\end{figure}

The methodology followed in this work consists of four main stages: the derivation of scenarios to be captured, the analysis of existing dataset, the derivation of ODDs from selected operational scenarios, the analysis of missing aspects and the planning of future data acquisition. Figure~\ref{fig:methodology} provides an overview of the overall process, which is described in detail in the following subsections.

\subsection{Dataset metrics}

\newcolumntype{C}[1]{>{\centering\arraybackslash}m{#1}}

\begin{table}[htb]
\renewcommand{\arraystretch}{1.3}
\caption{Evaluation of the information layers of the six layer model (top) and the additional categories proposed by Mendiboure et al. (bottom) with respect to dataset acquisition planning. Symbols denote low (--), medium (o), high (+), and very high (++) availability in existing datasets and ease of targeted acquisition. External data sources that can support the recording process are listed in the third column.} \label{tab:layers}
\centering
\begin{tabular}{|m{4.8cm}||
                C{2.8cm}|
                C{2.8cm}|
                C{3cm}|}
\hline
\textbf{Layer} & \textbf{Accessibility in Datasets} & \textbf{Planning for Recording} & \textbf{External Resources} \\ 
\hline \hline
Digital Information                         & --    & +     & -\\                \hline
Environmental Conditions                    & o     & o     & Open-Meteo \\     \hline
Dynamic Objects                             & ++    & --    & -\\                \hline
Temporary Modifications of L1 and L2        & --    & --    & -\\                \hline
Roadside Structures                         & o     & +     & OpenStreetMap \\  \hline
Road Network and Traffic Guidance Objects   & +     & ++    & OpenStreetMap \\  \hline \hline
Vehicle Information                         & ++    & ++    & -\\                \hline
Layer Scenarios                             & +     & --    & -\\                \hline
\end{tabular}
\end{table}



The accessibility of the different ODD elements in existing datasets and their suitability for planning future recordings is evaluated as shown in Table~\ref{tab:layers}. For most of the datasets, the recording vehicle is specified and the sensor setup is known, meaning that this information is directly available for acquisition. While information about Roadside Structures or the Road Network might be present in datasets, this can be extracted using the location of the recording and OpenStreetMap. Based on this information the acquisition can also be planned. Temporary Modifications are rarely explicitly annotated and are hard to plan for as they are not findable in a database similar to OpenStreetMap. While dynamic objects are the focus of most perception datasets and therefore are well annotated, their presence and distribution can't be anticipated for a recording. The exact environmental conditions are also rarely annotated, at best a general indication is given. A rough estimate can be retrieved from Open-Meteo. For planning recordings the general weather can be taken into account, but also a detailed planning is difficult. Digital information, unless already annotated or directly captured, is not possible to acquire afterwards. For the planning stage most information can be known before. Overall Roadside Structures, Road Network and Vehicle Information are most promising to both gather information about existing datasets as well as to plan for new recordings.

\subsection{Dataset Requirements and Acquisition Planning}

As a first step towards the creation of new datasets, it is necessary to define the scenarios that should be captured. Based on representative use cases, individual scenarios are derived and decomposed into their constituent elements according to the six layer model, vehicle-specific characteristics, and the interactions between road users. The resulting set of requirements serves as a reference for evaluating the coverage of existing datasets and should therefore be as comprehensive as possible.

The analysis of existing datasets focuses on identifying which of these elements are already represented and which remain underrepresented. In particular, dynamic objects, roadside structures, road network characteristics, vehicle information, and scenario descriptions can be readily extracted from existing datasets and therefore provide a basis for comparison. This analysis enables the identification of dataset gaps and the prioritization of scenarios that should be targeted in future recordings.

Based on the identified gaps, an acquisition strategy can be defined. Layers 1 and 2 of the six layer model are largely represented in OpenStreetMap and can therefore be used to systematically identify candidate recording locations. Examples include roads with cobblestone surfaces, narrow intersections, height restrictions, or tram crossings. In contrast, temporary infrastructure modifications and dynamic traffic interactions are difficult to plan in advance and must be captured opportunistically. Environmental conditions can be partially targeted using weather services such as Open-Meteo, while vehicle-specific information can be directly controlled during recording. This process enables the systematic selection of recording locations and conditions that maximize the coverage of underrepresented scenarios.

Following data acquisition, the recorded data are evaluated with respect to the coverage of the previously defined ODDs. Any ODD elements that remain insufficiently represented are identified as open requirements for future recording campaigns. This iterative process is repeated until adequate coverage of all targeted ODDs has been achieved.

\let\clearpage\relax
\section{Derivation of Dataset ODD from Scenarios}
\label{chapter:odd}
As introduced in Chapter~\ref{chapter:methodology}, the first step is to define the target ODD to be represented in the dataset. This ODD forms the basis for analyzing existing datasets and finding gaps between defined ODD and the recorded ODD. To achieve an ODD definition we introduce relevant scenarios for trucks as well as trams in urban environments and extract necessary entities to form the ODD.

\subsection{Truck ODD}
The ODD and use cases described by Martinez et al.~\cite{martinez_systematic_2025} are used as a reference for defining the scope of the truck dataset. These include turning maneuvers at intersections, side-by-side driving with other road users, reversing, and interactions with static obstacles, including both stationary objects and road infrastructure elements.

Due to their dimensions and limited maneuverability, trucks are particularly affected by scenarios in which the available space is constrained. Roundabouts represent a challenging example, especially in urban environments where their layout is often compact. During navigation of roundabouts, the maneuvering margin is reduced, increasing the risk of collisions with curbs, roadside infrastructure, or other road users. Similarly, intersections formed by narrow streets require precise vehicle control and are characterized by limited maneuvering space. Tight turning maneuvers are of particular interest, as the vehicle and trailer follow different trajectories and may require the occupation of adjacent road space to complete the turn safely.

Additional infrastructure-related scenarios should also be considered. Bridges, underpasses, and overhanging vegetation are relevant due to the height restrictions imposed on large vehicles and are largely absent from current perception datasets. Construction sites represent another important use case, as temporary barriers and lane modifications frequently reduce the available road width and create unusual traffic configurations.

Regarding side-by-side driving, narrow urban streets can create hazardous situations when trucks encounter large oncoming vehicles such as buses, trams, or other trucks. In such cases, the available lateral clearance may be insufficient, increasing the risk of contact between vehicle bodies or mirrors.

All types of road users are considered in the analysis, with particular emphasis on pedestrians and cyclists, as they are the most vulnerable participants in urban traffic. Regarding vehicle dynamics, the operating speed is assumed to remain below the urban speed limit of 50~km/h. However, since many challenging urban situations occur during low-speed operation, particular attention is given to scenarios involving speeds below 30~km/h and especially below 10~km/h.

For this reason, roundabouts and narrow intersections are analyzed as relevant infrastructure elements for truck operation. In addition, special attention is given to road segments shared by trucks and trams, as well as intersections containing tram tracks, as these scenarios represent potential interaction points between both modes of transport.

\subsection{Tram ODD}
Even though trams share the same environment as trucks and other road vehicles, due to their track bound nature they exhibit unique simplifications for automation as well as certain hindrances as described by Ruppert~\cite{ruppert_silent_2024}.
Mitusch et al.~\cite{mitusch_automatisierte_2023} categorize scenarios into three categories, simple, complex and highly complex. Simple scenarios are in areas where the tram doesn't interact with other road users due to a separate track body for example. As trucks are not able to traverse these areas we have only an overlap in ODD due to the sensor positions. Complex scenarios are located for example where the tracks are road bound. Here the tram is interacting with few other road users. In these areas trucks and trams share many entities in their specific ODDs. The main difference lies in the track bound nature of trams, besides that most entities are the same. This enables recordings in these areas to equally benefit both trucks and trams. Highly complex scenarios are located in areas with many road users as for example pedestrian zones. Even though trucks also operate in those areas data acquisition is complicated as these areas are often not open to public traffic, whereas trams operate here daily. As the focus of this work is on urban datasets in general we will focus on scenarios that trucks encounter, since they reach a larger area within the urban scope and take into account the shared ODD with trams. Therefore we will use scenarios that are relevant for trams, but accessible to trucks. 

A shared use case is movement along a street with embedded rails. Even if no tram is available as a recording vehicle it is possible to capture the sensor perspective and the relevant surroundings. Due to the difference in vehicle capabilities, the trajectory between both vehicles differs and also scenarios that depend on interaction with the ego-vehicle will be different. 

Other relevant use cases are interactions at tram level crossings. Since accidents in tram traffic often happen at interaction points, such as crossings, data at these points is both important for trams as well as other vehicles. For trams the challenge lies in anticipating road user behavior and for other road users in detecting trams approaching unsignaled crossings. 

These two use cases can be broken down into more finely granulated scenarios, but as to show the methodology of this work and to capture a diverse set of data it is kept broad at this point.

\subsection{Joint ODD Entities}
\begin{figure}[hptb]
    \centering{\includegraphics[width=8cm]{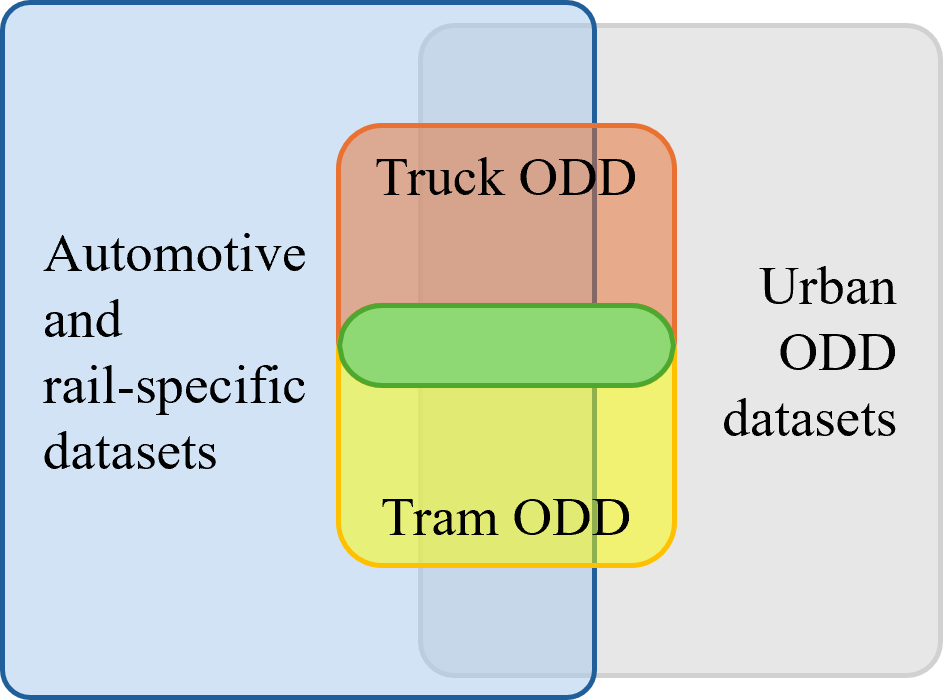}}
    \caption{Relationship between the coverage of existing automotive and rail-specific datasets, the urban ODD, and the proposed truck and tram ODDs. The green region highlights scenarios that are relevant to both truck and tram operation and therefore represent the most beneficial targets for future data acquisition.}
    \label{fig:odd_scope}
\end{figure}

As shown in Figure~\ref{fig:odd_scope}  currently existing datasets only partially cover the ODD of urban driving for all vehicles present in this environment. For our proposed example vehicles, heavy-duty truck and tram, certain aspects of existing datasets cover their ODDs, for example camera perception algorithms which are already generalized. Due to their dimensions and possible sensor mounting locations heavy-duty trucks and trams also share certain parts of their ODDs, therefore one dataset also enables a wider ODD for the other as shown in the subsection before.

By using the proposed method it is possible to find gaps in existing datasets and close them in a structured manner, taking into account vehicle particularities, capabilities and area of operation. 

\let\clearpage\relax
\section{Related work} \label{related_work}
This section reviews related work relevant to the scope of this study. For this, publicly available perception datasets are discussed, with a focus on urban environments. This analysis includes both automotive and rail-specific datasets.

\subsection{Automotive datasets}

\newcommand{\colWidthAuto}{3cm}

\begin{table}[htbp]
\renewcommand{\arraystretch}{1.3}
\caption{Comparison of the most relevant automotive perception datasets considered for this work. With respect to the NVIDIA dataset, zero classes were considered since these are not available at the time this paper is being written.}
\label{table:automotive_datasets}
\centering
\begin{tabular}{|p{1,5cm}||
                >{\centering\arraybackslash}m{\colWidthAuto}|
                >{\centering\arraybackslash}m{\colWidthAuto}|
                >{\centering\arraybackslash}m{\colWidthAuto}|
                >{\centering\arraybackslash}m{\colWidthAuto}|}
\hline
\textbf & \textbf{nuScenes \cite{caesar_nuscenes_2020}} 
        & \textbf{NVIDIA Physical AI \cite{nvidia_nvidia_2025}}
        & \textbf{TruckScenes \cite{fent_man_2024} }
        & \textbf{TruckDrive} \cite{ghilotti2026truckdrive} \\
\hline \hline
Scenes         & 1000     & 300k        & 747                   & 3828          \\ \hline
Scope       & Urban     & Urban  & Highway, Urban, Terminal    & Highway        \\ \hline
Region         & US, Asia & Europe, US  & Europe                & US            \\ \hline \hline
Camera         & 6        & 7           & 4                     & 11            \\ \hline
LiDAR          & 1        & 1           & 6                     & 10            \\ \hline
Radar          & 5        & 10          & 6                     & 10            \\ \hline
GNSS           & 1        & 1           & 1                     & 2             \\ \hline
IMU            & 1        & 1           & 2                     & 4             \\ \hline \hline
Classes        & 23       & 0*          & 27                    & 9             \\ \hline
Vehicle        & Car      & Car         & Truck                 & Truck         \\ \hline
\end{tabular}
\end{table}

Despite the lack of more detailed information, Liu et al.~\cite{liu_survey_2024} surveyed in 2024 265 different driving datasets, indicating that the number as of today may be larger. According to this work, only 48 datasets from the 191 analyzed are recorded in Europe, half of them in Germany, in contrast to 40 datasets recorded in the USA. This number indicates the regional heterogeneity among datasets. With regard to the scope of our work, these numbers may include datasets with a different sensor modality, with outdated sensor technologies, or with a non-urban scope. However, there is no more information apart from the publishing year. While Bogdoll et al.~\cite{bogdoll_ad-datasets_2022} included in 2022 more detailed information about more than 200 datasets, such as the sensor modality, regional information or publication year are missing.

To analyze relevant datasets, only urban datasets published in 2020 or later are considered, as earlier datasets may contain outdated information. Furthermore, the analysis is restricted to multimodal datasets due to the importance of sensor fusion in urban driving scenarios. Datasets are additionally required to include positioning sensors in order to enable the extraction of metadata related to the recorded road infrastructure. Finally, only datasets with a total recording duration exceeding one hour are included, in order to exclude small-scale datasets. While the geographical region of each dataset is taken into account, important datasets collected outside Europe are also included in this analysis.

Despite being recorded in the United States, the nuScenes dataset~\cite{caesar_nuscenes_2020}, published in 2020, is widely regarded as one of the most influential autonomous driving datasets in recent years. It comprises urban scenes from the cities of Boston and Singapore, covering multiple times of day and weather conditions, and includes high-definition map data. The sensor suite consists of cameras, LiDAR, and radar.

In 2025, NVIDIA released a large-scale dataset~\cite{nvidia_nvidia_2025} comprising approximately 1,700 hours of driving data collected under diverse weather and illumination conditions. The dataset includes scenarios recorded in both European and North American environments from the perspective of camera, LiDAR and radar sensors.

Although its primary focus is hub-to-hub operation, TruckScenes~\cite{fent_man_2024}, released in 2025, represents the first publicly available dataset recorded using a truck as research vehicle. In addition to highway driving, it includes a limited number of urban and terminal scenarios and is therefore considered relevant for this analysis. The used sensors are camera, LiDAR and radar. Also targeting truck automation, the recently published TruckDrive dataset~\cite{ghilotti2026truckdrive} was released in 2026. In contrast to TruckScenes, it focuses on American truck operations. The dataset features an extended sensor setup with an emphasis on long-range perception, making it particularly suitable for highway automation use cases. Table~\ref{table:automotive_datasets} includes an overview of the mentioned datasets.

Regarding city busses, to the best of our knowledge no publicly available large-scale dataset could be found for the research conducted.

\subsection{Rail-specific datasets}

\newcommand{\colWidthRail}{3cm}

\begin{table}[htbp]
\renewcommand{\arraystretch}{1.3}
\caption{Comparison of publicly available rail-specific perception datasets.} \centering
\label{table:rail_datasets}
\centering
\begin{tabular}{|p{1,5cm}||
                >{\centering\arraybackslash}m{\colWidthRail}|
                >{\centering\arraybackslash}m{\colWidthRail}|
                >{\centering\arraybackslash}m{\colWidthRail}|
                >{\centering\arraybackslash}m{\colWidthRail}|}
\hline
\textbf & \textbf{RailSem19 \cite{zendel_railsem19_2019}} 
        & \textbf{OSDaR23 \cite{tagiew_osdar23_2023}} 
        & \textbf{RailGoerl24 \cite{tagiew_railgoerl24_2025}} 
        & \textbf{Hovanotayan et al. \cite{hovanotayan_towards_2025}}  \\ 
        \hline \hline
Scenes         & 8500           & 45       & 12.2 k  & NA     \\ \hline
Scope    & Urban, regular operation  & Regular operation  & Closed-loop & Urban  \\ \hline
Region         & Worldwide      & Europe   & Europe  & Asia   \\ \hline \hline
Camera         & 1              & 9        & 1       & 1      \\ \hline
LiDAR          & 0              & 6        & 0       & 1      \\ \hline
Radar          & 0              & 1        & 1       & 0      \\ \hline
GNSS           & 0              & 1        & 0       & 1      \\ \hline
IMU            & 0              & 1        & 0       & 0      \\ \hline \hline
Classes        & 20             & 20       & 1       & 1      \\ \hline
Vehicle        & Train, Tram    & Train    & Train   & Tram   \\ \hline

\end{tabular}
\end{table}

With respect to rail-specific platforms, existing datasets have so far mostly focused on trains as data collection vehicles. Since the environment in which trains operate is usually closed and secure, perception tasks are not as extended as with automotive datasets. However, multiple datasets can be found in the literature. One of the largest datasets is the Open Sensor Data for Rails (OSDaR23) dataset~\cite{tagiew_osdar23_2023} released in 2023 in collaboration with the German railway operator Deutsche Bahn. This dataset targets train-based scenarios during the common operation of a line train. 20 different classes, including rail infrastructure and road users are labeled. Targeting road user safety, in the Görlitz Rail Test Center CV Dataset (RailGoerl24)~\cite{tagiew_railgoerl24_2025} different safety-critical scenarios were collected in a controlled and closed test environment, enabling a reproducible evaluation of perception system under challenging conditions. 

Regarding datasets that use a tram as research vehicle, the RailSem19 dataset~\cite{zendel_railsem19_2019} released in 2019 includes the segmented images from multiple videos  from the Internet under diverse situations in multiple weather and lighting conditions. 20 classes were annotated in total, including traffic infrastructure and road users. More recently, Hovanotayan et al.~\cite{hovanotayan_towards_2025} presented a tram-based dataset focusing on urban environments. To the best of our knowledge, this work represents the first publicly available dataset specifically targeting tram operation in complex urban traffic scenarios.

Despite some datasets being available to the public, it can be observed that the sensor modalities are usually limited and the data can hardly be used to train complex perception algorithms. Table~\ref{table:rail_datasets} includes an overview of the analyzed datasets.

\subsection{Extraction of dataset information}
With the goal of exemplifying the existing research gap, this chapter compares the ODD of trucks with the existing ODD of the datasets of nuScenes \cite{caesar_nuscenes_2020}, recorded in Boston and Singapore, and TruckScenes \cite{fent_man_2024}, recorded in the Bavaria region. The first dataset was used for analysis due to its relevancy and size, being used as benchmark for multiple approaches. The second one was selected because of using a truck as a recording vehicle, the recordings are also based in Germany, and its newness.

\begin{figure}[hptb]
    \centering{\includegraphics[width=1\textwidth]{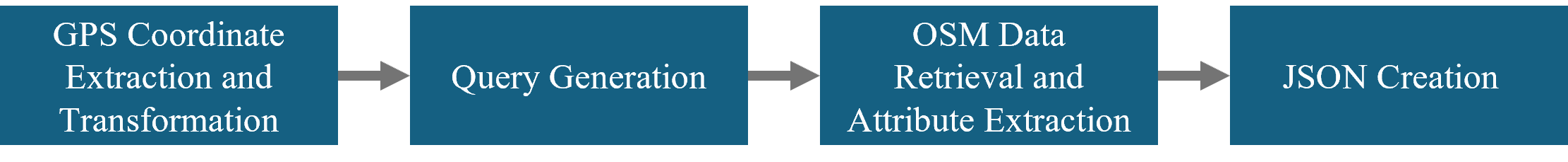}}
    \caption{Pipeline for extracting road infrastructure metadata from driving datasets. GPS coordinates are extracted from the dataset recordings, normalized to a common format, queried against OpenStreetMap via Overpass, and converted into structured JSON metadata.}
    \label{fig:dataset_pipeline}
\end{figure}

To identify the physical infrastructure represented in the analyzed datasets, metadata is extracted from OpenStreetMap along the recorded trajectories. The same procedure is applied to all datasets. First, the GPS coordinates are extracted and converted into a common format to ensure consistency across datasets. Subsequently, Overpass queries are generated based on these coordinates and executed against the OSM database. These queries retrieve information from nodes, ways, and relations located in the surroundings of the recorded positions. The returned OSM data includes infrastructure-related attributes such as road width, maximum permitted vehicle height, road type, or road configuration. The extracted information is then stored in JSON files, which are subsequently used to analyze and compare the infrastructure coverage of the considered datasets. An overview of the proposed pipeline is shown in Figure~\ref{fig:dataset_pipeline}.

\let\clearpage\relax
\section{Dataset acquisition planning based on identified gaps}

\begin{figure}[htb]
    \centering{\includegraphics[width=1\textwidth]{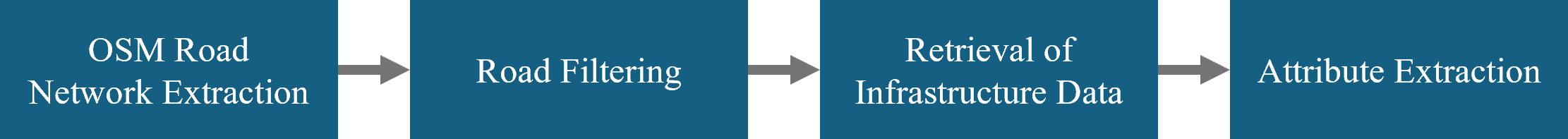}}
    \caption{Pipeline for extracting infrastructure metadata from the Darmstadt road network. Roads that are inaccessible to trucks and road types outside the scope of the proposed datasets are filtered out before querying OpenStreetMap. Relevant infrastructure attributes are subsequently extracted}
    \label{fig:da_pipeline}
\end{figure}

Since the target region of the proposed datasets is Darmstadt and its surrounding area, a second processing pipeline is used to extract infrastructure information directly from the region of interest. First, road segments that are inaccessible to trucks are filtered out. Subsequently, road types that are outside the scope of the intended datasets, such as \textit{footway}, \textit{pedestrian}, or \textit{proposed} roads, are excluded from the analysis. The remaining road network is then queried using the OSM database to extract the same set of infrastructure-related attributes considered in the dataset analysis. This process provides a representation of the infrastructure characteristics present in the target area and enables a comparison with the coverage of existing datasets. An overview of the proposed pipeline is shown in Figure~\ref{fig:da_pipeline}.

The search radius was set to 10~km, with Luisenplatz selected as the center point. Luisenplatz is located in the center of Darmstadt and constitutes one of the city's main transportation and commuting hubs. The selected radius was chosen to cover not only the urban core but also the surrounding districts and industrial areas that are relevant for truck and tram operation. A smaller radius would exclude important peripheral infrastructure and delivery locations, whereas a significantly larger radius would include predominantly suburban and rural road networks that fall outside the intended urban ODD. Within this area, a total of 19,981 unique road segments (\textit{ways}) were identified. The analysis focused on road infrastructure elements relevant to truck operation, including roundabouts and intersection widths, as well as tram crossings, which are particularly important in urban environments where road and rail traffic interact. 

\subsection{Roundabouts}
Roundabouts represent a particularly relevant infrastructure element for truck operation due to the limited maneuvering space available to large vehicles. To identify roundabouts accessible to trucks within the area of interest, the OSM query shown below was executed. The query selects all road segments tagged as \texttt{junction=roundabout} while excluding pedestrian, bicycle, and other non-drivable infrastructure. In addition, road segments with restrictions preventing truck access were filtered out.

\begin{tcolorbox}
\begin{verbatim}
[out][timeout:100];
(
way(around:10000,49.87290763602407,8.651291718617319)
[junction=roundabout]
[highway]
["highway"!~"^(footway|cycleway|path|steps|pedestrian|bridleway|
corridor|elevator|escalator|platform|construction|proposed)$"]
["hgv"!~"^(no|private|destination|delivery|
agricultural|forestry)$"];
);
out geom;
\end{verbatim}
\end{tcolorbox}

The query returned 66 road segments tagged as roundabouts. Since a single roundabout may be represented by multiple connected ways in OSM, a post-processing step was applied to merge duplicate representations and identify unique roundabout structures. After this cleaning process, a total of 33 unique roundabouts were identified within the analyzed area. These roundabouts constitute relevant candidate locations for future data acquisition, as they represent challenging maneuvering scenarios that are particularly important for truck operation. For comparison, only four roundabouts were identified in the nuScenes dataset (three in Boston and one in Singapore), while TruckScenes contains ten roundabouts. Consequently, the target area offers substantially greater coverage of this infrastructure element and provides opportunities to enrich future datasets with a wider variety of roundabout configurations.


\subsection{Narrow Intersections}
Narrow intersections represent a particularly challenging scenario for trucks, especially when turning from one narrow road into another. Due to the vehicle dimensions and turning radius, such intersections may offer limited maneuvering space and increase the likelihood of interactions with roadside infrastructure, parked vehicles, or other road users. To identify potentially relevant scenarios, only intersections for which both connecting roads have a defined \texttt{width} attribute were considered. When multiple road segments belonged to the same road group, a representative width was determined using the median of the available width values. The resulting distribution is shown in Figure~5.

\begin{figure}[htb]
    \centering{\includegraphics[width=0.8\linewidth]{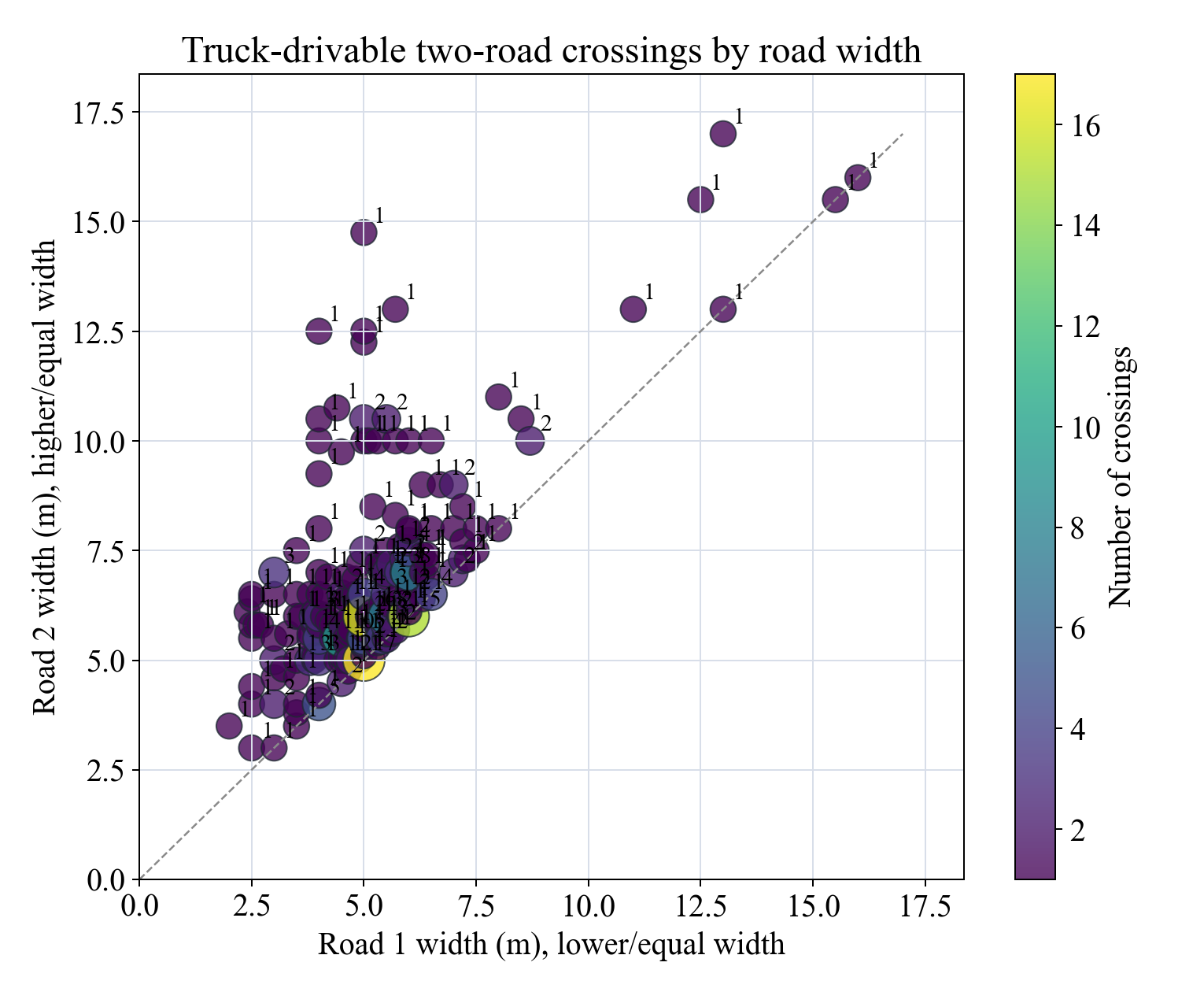}}
    \caption{Distribution of intersection road widths in the analyzed area. Each point represents a pair of intersecting roads, grouped according to their representative widths.}
    \label{fig:xy_roadwidth}
\end{figure}

For each valid intersection, the two road widths were sorted such that
\[ w_1 \leq w_2, \]
thereby ensuring that an intersection between roads with widths of \(2\,\mathrm{m}\) and \(3\,\mathrm{m}\) was treated identically to one between roads with widths of \(3\,\mathrm{m}\) and \(2\,\mathrm{m}\). The resulting width pairs \((w_1, w_2)\) were then aggregated by counting the number of intersections associated with each pair.

As shown in Figure~5, the majority of intersections involve roads with sufficient width for truck operation. Nevertheless, a number of intersections were identified in which both roads have widths close to \(2.5\,\mathrm{m}\). Given that the maximum legal vehicle width in Europe is approximately \(2.55\,\mathrm{m}\), these locations may constitute challenging maneuvering scenarios, particularly during turning maneuvers. Consequently, such intersections should be considered candidate locations for future data acquisition.

\subsection{Tram crossings}
Candidate crossing nodes were identified using the tag \texttt{railway=tram\_level\_crossing}. For each candidate node, the corresponding parent ways were retrieved and classified into two categories: road ways tagged as \texttt{highway} and rail ways tagged as either \texttt{railway=tram} or \texttt{railway=light\_rail}. To retain only crossings relevant to truck operation, the parent road ways were filtered according to their road classification and access restrictions, following the methodology described in the previous sections. Candidate nodes without at least one remaining truck-accessible road were subsequently discarded.

\begin{figure}[htb]
    \centering
    \includegraphics[width=0.7\linewidth]{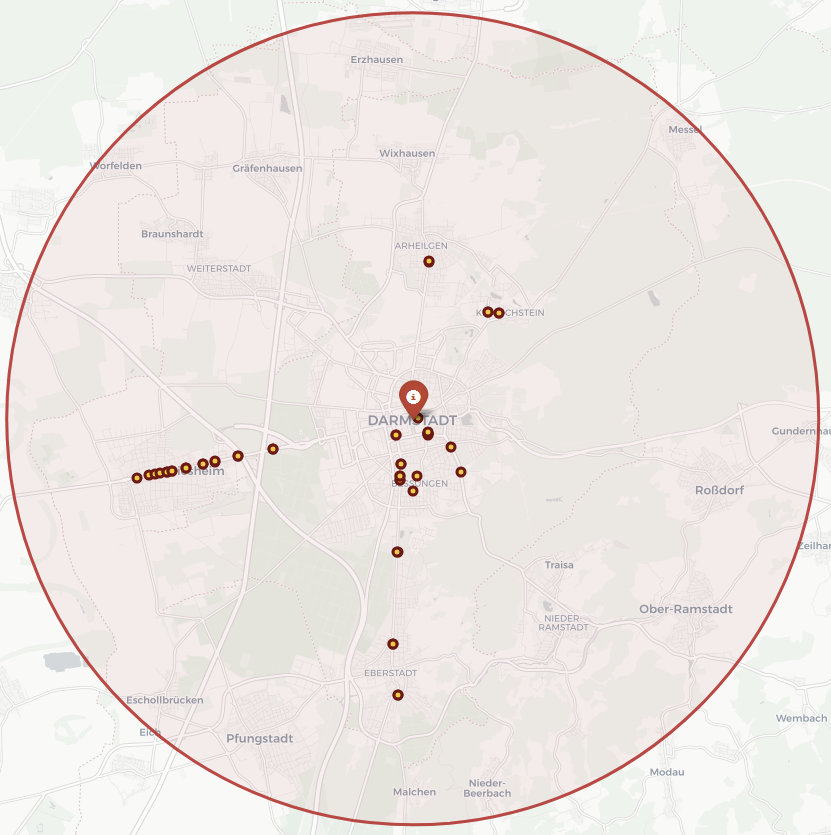}
    \caption{Spatial distribution of intersection candidates within the analyzed area around Darmstadt. The red marker indicates the center point at Luisenplatz, and the red circle denotes the search radius. Yellow points mark the identified intersection locations.}
    \label{fig:tram-crossings}
\end{figure}

The search identified a total of 49 tram crossings, whose spatial distribution is shown in Figure~6. Most of these crossings are located in the municipality of Griesheim, highlighting this area as a particularly relevant environment for studying interactions between road and tram traffic.

\subsection{Truck Dataset}

Based on the findings of this work and the analysis of additional infrastructure elements, selected scenes are recorded using a truck as the research vehicle. The vehicle and its sensor setup are presented in~\cite{martinez_systematic_2025}. The recordings cover the identified points of interest and include positional data, camera images, point clouds, and driver prompts, with a future perspective towards end-to-end driving research. A sample of the data collected using the methodology presented in this paper is available at~\cite{martinezDataset}.

\section{Conclusion and Outlook}
This paper highlights the imbalance present in autonomous driving research and datasets between passenger vehicles and other platforms such as trucks, buses, and trams. This gap is considered critical due to the increasing driver shortage and the growing demand for automated solutions in logistics and public transportation.

Following the analysis of multiple automotive and rail-specific datasets, the need for urban datasets representing a broader range of vehicle platforms was identified. The results show that current datasets provide limited coverage of scenarios that are particularly relevant for truck and tram operation in urban environments. Consequently, they are insufficient for the development and fine-tuning of perception and decision-making approaches targeting these platforms.

To address this limitation, a systematic methodology was proposed for deriving dataset requirements from Operational Design Domains and representative use cases. Furthermore, the target acquisition area was analyzed using OpenStreetMap data. This analysis identified several infrastructure elements of interest, including 33 truck-accessible roundabouts, narrow intersections with limited maneuvering space, and 49 tram crossings. These results demonstrate the availability of numerous candidate locations for future data acquisition and provide a quantitative basis for planning recordings that address the identified dataset gaps.

By clearly defining the problem and deriving representative use cases for future truck and tram datasets, this work establishes a structured approach for the creation of new datasets that complement existing resources and expand their coverage of urban scenarios.

As a next step, this work will be extended through the creation of a dedicated tram dataset and the expansion of the presented truck dataset. Both datasets will target real-world urban scenarios derived from the previously defined ODDs and use cases, with particular emphasis on challenging and safety-critical situations that remain underrepresented in existing datasets.

These datasets are intended to complement large, general-purpose autonomous driving datasets by enabling platform-specific fine-tuning and validation. Ultimately, they aim to support the development, evaluation, and comparison of perception and automation algorithms for heavy-duty and rail-based vehicles.

\printbibliography

@inproceedings{caesar_nuscenes_2020,
	title = {{nuScenes}: A Multimodal Dataset for Autonomous Driving},
	issn = {2575-7075},
	url = {https://ieeexplore.ieee.org/document/9156412},
	doi = {10.1109/CVPR42600.2020.01164},
	shorttitle = {{nuScenes}},
	eventtitle = {2020 {IEEE}/{CVF} Conference on Computer Vision and Pattern Recognition ({CVPR})},
	pages = {11618--11628},
	booktitle = {2020 {IEEE}/{CVF} Conference on Computer Vision and Pattern Recognition ({CVPR})},
	author = {Caesar, Holger and Bankiti, Varun and Lang, Alex H. and Vora, Sourabh and Liong, Venice Erin and Xu, Qiang and Krishnan, Anush and Pan, Yu and Baldan, Giancarlo and Beijbom, Oscar},
	urldate = {2026-01-27},
	date = {2020-06},
	note = {{ISSN}: 2575-7075},
}

@article{liu_survey_2024,
	title = {A Survey on Autonomous Driving Datasets: Statistics, Annotation Quality, and a Future Outlook},
	volume = {9},
	issn = {2379-8904},
	url = {https://ieeexplore.ieee.org/document/10509812},
	doi = {10.1109/TIV.2024.3394735},
	shorttitle = {A Survey on Autonomous Driving Datasets},
	pages = {7138--7164},
	number = {11},
	journaltitle = {{IEEE} Transactions on Intelligent Vehicles},
	author = {Liu, Mingyu and Yurtsever, Ekim and Fossaert, Jonathan and Zhou, Xingcheng and Zimmer, Walter and Cui, Yuning and Zagar, Bare Luka and Knoll, Alois C.},
	urldate = {2026-01-27},
	date = {2024-11},
}

@inproceedings{fent_man_2024,
	location = {Vancouver, {BC}, Canada},
	title = {{MAN} {TruckScenes}: A multimodal dataset for autonomous trucking in diverse conditions},
	isbn = {979-8-3313-1438-5},
	url = {http://www.proceedings.com/079017-1982.html},
	doi = {10.52202/079017-1982},
	shorttitle = {{MAN} {TruckScenes}},
	eventtitle = {Advances in Neural Information Processing Systems 37},
	pages = {62062--62082},
	booktitle = {Advances in Neural Information Processing Systems 37},
	publisher = {Neural Information Processing Systems Foundation, Inc. ({NeurIPS})},
	author = {Fent, Felix and Juergens, Stefan and Kuttenreich, Fabian and Lechermann, Lorenz and Lienkamp, Markus and Nissler, Christian and Perl, Andrea and Rizwin, Farija and Ruch, Florian and Voll, Ulrich and Yan, Min},
	urldate = {2026-01-27},
	date = {2024},
	langid = {english},
}

@inproceedings{hovanotayan_towards_2025,
	title = {Towards Pedestrian Trajectory Prediction in Tram Environments for Safe and Efficient Autonomous Tram Systems: Dataset Development},
	url = {https://ieeexplore.ieee.org/abstract/document/11216747},
	doi = {10.1109/ICIRT66379.2025.11216747},
	shorttitle = {Towards Pedestrian Trajectory Prediction in Tram Environments for Safe and Efficient Autonomous Tram Systems},
	eventtitle = {2025 {IEEE} International Conference on Intelligent Rail Transportation ({ICIRT})},
	pages = {74--79},
	booktitle = {2025 {IEEE} International Conference on Intelligent Rail Transportation ({ICIRT})},
	author = {Hovanotayan, Supatat and Nakano, Kimihiko and Takata, Tetsuya and Nagasawa, Hiroyuki and Ueda, Makoto and Sakakidani, Kazuya},
	urldate = {2026-01-26},
	date = {2025-10},
	note = {Tram Dataset},
}

@inproceedings{zendel_railsem19_2019,
	title = {{RailSem}19: A Dataset for Semantic Rail Scene Understanding},
	url = {https://openaccess.thecvf.com/content_CVPRW_2019/html/WAD/Zendel_RailSem19_A_Dataset_for_Semantic_Rail_Scene_Understanding_CVPRW_2019_paper.html},
	shorttitle = {{RailSem}19},
	eventtitle = {Proceedings of the {IEEE}/{CVF} Conference on Computer Vision and Pattern Recognition Workshops},
	author = {Zendel, Oliver and Murschitz, Markus and Zeilinger, Marcel and Steininger, Daniel and Abbasi, Sara and Beleznai, Csaba},
	urldate = {2026-01-26},
	date = {2019},
	note = {Tram Dataset},
}

@inproceedings{tagiew_osdar23_2023,
	title = {{OSDaR}23: Open Sensor Data for Rail 2023},
	url = {http://arxiv.org/abs/2305.03001},
	doi = {10.1109/ICRAE59816.2023.10458449},
	shorttitle = {{OSDaR}23},
	pages = {270--276},
	booktitle = {2023 8th International Conference on Robotics and Automation Engineering ({ICRAE})},
	author = {Tagiew, Rustam and Köppel, Martin and Schwalbe, Karsten and Denzler, Patrick and Neumaier, Philipp and Klockau, Tobias and Boekhoff, Martin and Klasek, Pavel and Tilly, Roman},
	urldate = {2025-01-27},
	date = {2023-11-17},
	langid = {english},
	eprinttype = {arxiv},
	eprint = {2305.03001 [cs]},
}

@misc{tagiew_railgoerl24_2025,
	title = {{RailGoerl}24: Görlitz Rail Test Center {CV} Dataset 2024},
	url = {http://arxiv.org/abs/2504.00204},
	doi = {10.48550/arXiv.2504.00204},
	shorttitle = {{RailGoerl}24},
	number = {{arXiv}:2504.00204},
	publisher = {{arXiv}},
	author = {Tagiew, Rustam and Wunderlich, Ilkay and Sastuba, Mark and Seitz, Steffen},
	urldate = {2025-04-28},
	date = {2025-03-31},
	langid = {english},
	eprinttype = {arxiv},
	eprint = {2504.00204 [cs]},
}

@article{ji-hyland_what_2022,
	title = {What do professional drivers think about their profession? An examination of factors contributing to the driver shortage},
	volume = {25},
	issn = {1367-5567, 1469-848X},
	url = {https://www.tandfonline.com/doi/full/10.1080/13675567.2020.1821623},
	doi = {10.1080/13675567.2020.1821623},
	shorttitle = {What do professional drivers think about their profession?},
	pages = {231--246},
	number = {3},
	journaltitle = {International Journal of Logistics Research and Applications},
	shortjournal = {International Journal of Logistics Research and Applications},
	author = {Ji-Hyland, Chao and Allen, Declan},
	urldate = {2026-01-27},
	date = {2022-03-04},
	langid = {english},
}

@online{hr_wie_nodate,
	title = {Wie sich der Busfahrer-Mangel in Deutschland auswirkt},
	url = {https://www.tagesschau.de/wirtschaft/arbeitsmarkt/oepnv-busfahrer-nachwuchsmangel-100.html},
	titleaddon = {tagesschau.de},
	author = {hr, Katharina Bruns},
	urldate = {2026-01-27},
	langid = {german},
}

@online{nvidia_nvidia_2025,
	title = {{NVIDIA} Unveils Open Physical {AI} Dataset to Advance Robotics and Autonomous Vehicle Development},
	url = {https://blogs.nvidia.com/blog/open-physical-ai-dataset/},
	titleaddon = {{NVIDIA} Blog},
	author = {{NVIDIA}},
	urldate = {2026-01-27},
	date = {2025-03-18},
	langid = {american},
}

@inproceedings{mendiboure_operational_2023,
	title = {Operational Design Domain for Automated Driving Systems: Taxonomy Definition and Application},
	issn = {2642-7214},
	url = {https://ieeexplore.ieee.org/document/10186765},
	doi = {10.1109/IV55152.2023.10186765},
	shorttitle = {Operational Design Domain for Automated Driving Systems},
	eventtitle = {2023 {IEEE} Intelligent Vehicles Symposium ({IV})},
	pages = {1--6},
	booktitle = {2023 {IEEE} Intelligent Vehicles Symposium ({IV})},
	author = {Mendiboure, Leo and Benzagouta, Mohamed Lamine and Gruyer, Dominique and Sylla, Tidiane and Adedjouma, Morayo and Hedhli, Abdelmename},
	urldate = {2026-01-28},
	date = {2023-06},
}

@online{european_commission_etcs_nodate,
	title = {{ETCS} Levels and Modes - Mobility and Transport},
	url = {https://transport.ec.europa.eu/transport-modes/rail/ertms/what-ertms-and-how-does-it-work/etcs-levels-and-modes_en},
	author = {{European Commission}},
	urldate = {2026-01-28},
	langid = {english},
}

@online{torc_robotics_solutions_2023,
	title = {Solutions},
	url = {https://torc.ai/solutions/},
	author = {{Torc Robotics}},
	urldate = {2026-01-28},
	date = {2023-10-11},
	langid = {american},
}

@online{aurora_aurora_nodate,
	title = {Aurora: Self-driving freight is here.},
	url = {https://aurora.tech/},
	shorttitle = {Aurora},
	author = {{Aurora}},
	urldate = {2026-01-28},
}

@online{waymo_hello_nodate,
	title = {Hello London! Your Waymo ride is arriving},
	url = {https://waymo.com/blog/2025/10/hello-london-your-waymo-ride-is-arriving},
	titleaddon = {Waymo},
	author = {{Waymo}},
	urldate = {2026-01-28},
	langid = {english},
}

@article{schuster_will_2023,
	title = {Will automated vehicles solve the truck driver shortages? Perspectives from the trucking industry},
	volume = {74},
	doi = {10.1016/j.techsoc.2023.102313},
	shorttitle = {Will automated vehicles solve the truck driver shortages?},
	pages = {102313},
	journaltitle = {Technology in Society},
	shortjournal = {Technology in Society},
	author = {Schuster, Amy and Agrawal, Shubham and Britt, Noah and Sperry, Danielle and Van Fossen, Jenna and Wang, Sicheng and Mack, Elizabeth and Liberman, Jessica},
	date = {2023-07-01},
}

@inproceedings{bogdoll_ad-datasets_2022,
	location = {Online Streaming, --- Select a Country ---},
	title = {Ad-datasets: A Meta-collection of Data Sets for Autonomous Driving:},
	isbn = {978-989-758-573-9},
	url = {https://www.scitepress.org/DigitalLibrary/Link.aspx?doi=10.5220/0011001900003191},
	doi = {10.5220/0011001900003191},
	shorttitle = {Ad-datasets},
	eventtitle = {8th International Conference on Vehicle Technology and Intelligent Transport Systems},
	pages = {46--56},
	booktitle = {Proceedings of the 8th International Conference on Vehicle Technology and Intelligent Transport Systems},
	publisher = {{SCITEPRESS} - Science and Technology Publications},
	author = {Bogdoll, Daniel and Schreyer, Felix and Zöllner, J.},
	urldate = {2026-01-30},
	date = {2022},
	langid = {english},
}

@report{martinez_systematic_2025,
	title = {Systematic Derivation of Requirements for the Perception Task of Free Space Driving Assistance Function Applied to Trucks},
	issn = {0148-7191, 2688-3627},
	url = {https://legacy.sae.org/publications/technical-papers/content/2025-01-0279/},
	doi = {10.4271/2025-01-0279},
	number = {2025-01-0279},
	institution = {{SAE} Technical Paper},
	author = {Martinez, Cristian and Peters, Steven},
	urldate = {2026-05-29},
	date = {2025-07-02},
	doi = {10.4271/2025-01-0279},
}

@report{mitusch_automatisierte_2023,
	title = {Automatisierte Betriebsfunktionen von Straßenbahnen: Bewertung der Potenziale von Technologien zum vernetzten Fahren},
	url = {https://www.dzsf.bund.de/SharedDocs/Downloads/DZSF/Veroeffentlichungen/Forschungsberichte/2023/ForBe_38_2023_AutoTram.html},
	shorttitle = {Automatisierte Betriebsfunktionen von Straßenbahnen},
	pages = {259 Seiten, ca. 10 MB},
	institution = {Deutsches Zentrum für Schienenverkehrsforschung beim Eisenbahn-Bundesamt},
	author = {Mitusch, Kay and Szimba, Eckhard and Knoch, Eva-Maria and Knebel, Robin and Berthold, Kilian and Fesser, Jonas and Canales, Carlos and Frindik, Roland},
	editora = {Heizler, Felix},
	editoratype = {collaborator},
	urldate = {2025-01-21},
	date = {2023},
	langid = {german},
	doi = {10.48755/DZSF.230006.01},
	note = {Artwork Size: 259 Seiten, ca. 10 {MB}},
}

@thesis{ruppert_silent_2024,
	location = {Darmstadt},
	title = {Silent Testing der Fahrzeugsteuerung zur Identifikation von Herausforderungen und Möglichkeiten des Automatisierten Fahrens von Straßenbahnen},
	url = {https://tuprints.ulb.tu-darmstadt.de/26362},
	institution = {Technische Universität Darmstadt},
	type = {Dissertation},
	author = {Ruppert, Timm},
	urldate = {2026-05-30},
	date = {2024-05-08},
	langid = {german},
	doi = {10.26083/tuprints-00026362},
}

@article{scholtes_6-layer_2021,
	title = {6-Layer Model for a Structured Description and Categorization of Urban Traffic and Environment},
	volume = {9},
	rights = {https://creativecommons.org/licenses/by/4.0/legalcode},
	issn = {2169-3536},
	url = {https://ieeexplore.ieee.org/document/9400833/},
	doi = {10.1109/ACCESS.2021.3072739},
	pages = {59131--59147},
	journaltitle = {{IEEE} Access},
	shortjournal = {{IEEE} Access},
	author = {Scholtes, Maike and Westhofen, Lukas and Turner, Lara Ruth and Lotto, Katrin and Schuldes, Michael and Weber, Hendrik and Wagener, Nicolas and Neurohr, Christian and Bollmann, Martin Herbert and Kortke, Franziska and Hiller, Johannes and Hoss, Michael and Bock, Julian and Eckstein, Lutz},
	urldate = {2026-05-30},
	date = {2021},
}

@misc{OpenStreetMap,
   author = {{OpenStreetMap contributors}},
   title = {{Planet dump retrieved from https://planet.osm.org }},
   howpublished = "\url{ https://www.openstreetmap.org }",
   year = {2017},
 }

@software{Zippenfenig_Open-Meteo,
  author = {Zippenfenig, Patrick},
  doi = {10.5281/zenodo.7970649},
  licence = {CC-BY-4.0},
  title = {Open-Meteo.com Weather API},
  year = {2023},
  copyright = {Creative Commons Attribution 4.0 International},
  url = {https://open-meteo.com/}
}

@inproceedings{ghilotti2026truckdrive,
  author    = {Ghilotti, Filippo and Palladin, Edoardo and Brucker, Samuel and Sigal, Adam and Bijelic, Mario and Heide, Felix},
  title     = {TruckDrive: Long-Range Autonomous Highway Driving Dataset},
  booktitle = {Proceedings of the IEEE/CVF Conference on Computer Vision and Pattern Recognition (CVPR)},
  year      = {2026},
}

@misc{martinezDataset,
  title = {Dataset for {{Paper}} "{{Towards Safer Cities}}: {{A Design Process}} for {{Perception Datasets}} in {{Urban Driving Scenarios}}"},
  author = {Martinez, Cristian and Kroth, Alexander and Peters, Steven},
  year = 2026,
  month = aug,
  url = {https://tudatalib.ulb.tu-darmstadt.de/handle/tudatalib/5563},
  publisher = {TUdatalib},
  address = {Darmstadt, Germany}
}
\newpage

\end{document}